\documentclass[10pt,twocolumn,letterpaper]{article}

\usepackage{iccv}              

\definecolor{iccvblue}{rgb}{0.21,0.49,0.74}
\usepackage[pagebackref,breaklinks,colorlinks,allcolors=iccvblue]{hyperref}

\usepackage{times}
\usepackage{epsfig}
\usepackage{graphicx}
\usepackage{amsmath}
\usepackage{amssymb}
\usepackage{bm}
\usepackage{float}
\usepackage{tabularx}
\usepackage{microtype}

\usepackage{booktabs} 
\usepackage{url}
\usepackage{caption}
\usepackage{multirow}
\usepackage{multicol}
\usepackage{array}
\usepackage{pifont}
\usepackage{xcolor}
\usepackage{booktabs}
\usepackage[normalem]{ulem}
\usepackage{xspace}
\usepackage{diagbox}
\usepackage{color}
\usepackage{enumitem}
\usepackage{bm}
\usepackage{algorithm}
\usepackage{listings}
\usepackage{xspace}
\usepackage{makecell}

\newcommand{\tsb}{\textsubscript}
\newcommand{\method}{APT}
\newcommand\blfootnote[1]{\begingroup\renewcommand\thefootnote{}\footnote{#1}\addtocounter{footnote}{-1}\endgroup}

\def\paperID{3523} 
\def\confName{ICCV}
\def\confYear{2025}

\title{Achieving More with Less: Additive Prompt Tuning for Rehearsal-Free Class-Incremental Learning}

\author{%
  Haoran Chen$^{1,2}$~\hspace{5pt}
  Ping Wang$^{1,2}$~\hspace{5pt} 
  Zihan Zhou$^{1,2}$~\hspace{5pt} 
  Xu Zhang$^{3}$~\hspace{5pt} 
  Zuxuan Wu$^{1,2\dagger}$~\hspace{5pt}
  Yu-Gang Jiang$^{1,2}$~\hspace{5pt}
\vspace{0.1in}\\ 
$^1$Institute of Trustworthy Embodied AI, Fudan University \\
$^2$Shanghai Collaborative Innovation Center of Intelligent Visual Computing \\
$^{3}$APUS AI Lab
}

\begin{document}
\maketitle
\begin{abstract}
Class-incremental learning (CIL) enables models to learn new classes progressively while preserving knowledge of previously learned ones. Recent advances in this field have shifted towards parameter-efficient fine-tuning techniques, with many approaches building upon the framework that maintains a pool of learnable prompts. Although effective, these methods introduce substantial computational overhead, primarily due to prompt pool querying and increased input sequence lengths from prompt concatenation. In this work, we present a novel prompt-based approach that addresses this limitation. Our method trains a single set of shared prompts across all tasks and, rather than concatenating prompts to the input, directly modifies the CLS token's attention computation by adding the prompts to it. This simple and lightweight design not only significantly reduces computational complexity—both in terms of inference costs and the number of trainable parameters—but also eliminates the need to optimize prompt lengths for different downstream tasks, offering a more efficient yet powerful solution for rehearsal-free class-incremental learning. Extensive experiments across a diverse range of CIL benchmarks demonstrate the effectiveness of our approach, highlighting its potential to establish a new prompt-based CIL paradigm. Furthermore, experiments on general recognition benchmarks beyond the CIL setting also show strong performance, positioning our method as a promising candidate for a general parameter-efficient fine-tuning approach. Code is available at \href{https://github.com/HaoranChen/Additive-Prompt-Tuning}{https://github.com/HaoranChen/Additive-Prompt-Tuning}.

\blfootnote{$^{\dagger}$ Corresponding author.}
\end{abstract}    
\section{Introduction}
\label{sec:intro}
Class-incremental learning (CIL) is a paradigm in continual learning where a model learns new classes progressively while preserving knowledge of previously learned ones~\cite{de2021defysurvey, masana2022classsurvey, wang2024comprehensivesurvey}. Unlike traditional machine learning, where all classes are presented simultaneously during training, CIL introduces classes sequentially. One of the central challenges in CIL is mitigating catastrophic forgetting~\cite{french1999catastrophic, kemker2018measuring, kirkpatrick2017overcoming}, a phenomenon where the model's performance on earlier classes degrades as it adapts to new ones. 

To address this, traditional methods often rely on rehearsal-based approaches~\cite{rebuffi2017icarl, belouadah2019il2m}, which maintain a memory buffer containing a subset of data from previous tasks. This buffer is periodically used to retrain the model when new tasks are introduced. However, due to concerns such as data privacy, most recent CIL methods have increasingly shifted toward rehearsal-free scenarios~\cite{lomonaco2020rehearsalfree, smith2023closer, chen2024adaptive}. This has inspired research into leveraging parameter-efficient fine-tuning (PEFT)~\cite{xin2024parameter} techniques, particularly those that utilize visual prompt tokens introduced by VPT~\cite{jia2022visual}. A pioneering work in this direction is L2P~\cite{wang2022l2p}, which proposes a framework built on three key components: (1) maintaining a pool of learnable prompts, (2) employing a query-based mechanism to select a subset of relevant prompts for each input instance, and (3) prepending the selected prompts to the input embeddings for subsequent training. This design has inspired numerous works~\cite{wang2022dualprompt, smith2023coda, wang2023isolation, chen2024promptfusion} that adopt a similar pipeline while introducing specialized modifications, rapidly achieving leading scores across multiple continual learning benchmarks.

Nevertheless, this framework introduces several significant but often overlooked drawbacks. First, from a practical point of view, its query-based mechanism requires an additional forward pass through a raw Vision Transformer~\cite{dosovitskiy2020image} model, effectively doubling the computational cost. Furthermore, to achieve competitive performance, most methods choose to use a large number of prompt tokens, increasing the input sequence length by up to 50\%~\cite{wang2023isolation}. This in turn results in a proportional increase in computational overhead, as the computational cost of ViTs scales directly with the number of input tokens. Meanwhile, recent research~\cite{huang2024ovor} demonstrates that using a single prompt for all tasks can achieve competitive performance. This finding raises important questions: Is such a complex prompt selection framework truly necessary? And is the paradigm of concatenating prompt tokens the most effective and efficient approach for prompt-based continual learning?

To address these challenges, we propose a novel prompt-based paradigm for continual learning. Specifically, instead of relying on a prompt pool, we train a single set of prompt tokens shared across all tasks. This approach not only reduces computational cost but also eliminates the need for complex loss function designs typically required for prompt selection. Furthermore, we replace the traditional concatenation of prompts with additive operations, providing a simpler and more efficient mechanism to influence the model's feature extraction process. By doing so, the original sequence length is preserved, avoiding the computational overhead associated with extending the input through concatenated prompt tokens.

However, implementing prompt token addition introduces a potential challenge. If we naively choose to train a set of prompts and add them to every token in the sequence, the number of trainable prompt parameters would become prohibitively large. On the other hand, selectively adding to only a subset of tokens raises the complex question of which tokens to modify, a decision that could lead to tedious and impractical experimentation. To address this, we draw inspiration from recent research~\cite{walmer2023teaching,wang2024cls} suggesting that the CLS token effectively captures key visual information for input images. Based on this insight, we propose adding only to the CLS token at each transformer block. 

Moreover, instead of applying addition at the input level, we incorporate prompt addition directly into the keys and values generated by the CLS token during the self-attention process. This design ensures that the number of prompts per layer is fixed at 2. For a ViT-B/16 architecture with 12 transformer layers, our approach requires training only 24 prompt vectors. This is significantly more parameter-efficient compared to previous approaches that maintain large prompt pools, such as in Coda-Prompt where the prompt pool size is 100. Meanwhile, an additional benefit of this approach is that it eliminates the need to explicitly determine the optimal number of prompts for different downstream datasets, simplifying both implementation and optimization.

To further mitigate forgetting, we propose a Progressive Prompt Fusion (PPF) inference strategy for the prompts. Specifically, we save the original set of prompts before training on a new task and compute a weighted average of the original and newly trained prompts after completing training~\cite{cai2021exponential}. This simple strategy ensures that the prompts retain knowledge from previously learned tasks while incorporating relevant information from the current task, enabling a smooth and effective continual learning process.

We name our method \method~(\textbf{A}dditive \textbf{P}rompt \textbf{T}uning) and conduct extensive experiments on 4 popular CIL benchmarks, namely CIFAR-100, ImageNet-R, CUB200, and Stanford Cars, to show its effectiveness. The results demonstrate that \method~achieves state-of-the-art performance compared to other prompt-based methods, while significantly reducing inference costs and the number of trainable prompt parameters. For example, on ImageNet-R, one of the most challenging class-incremental learning datasets, our method can achieve an average performance gain of 5.2\% with 41.5\% GFLOPs reduction and 78.2\% less trainable prompt parameters. Furthermore, we evaluate our approach on a diverse range of downstream recognition tasks beyond the continual learning setting. Given that our method is a direct refinement of VPT~\cite{jia2022visual}, we primarily focus on comparisons with it. The results again demonstrate that \method~not only outperforms VPT but also achieves significant reductions in both inference time and computational overhead. In summary, our contributions are three-fold:
\begin{itemize}
    \item[$\bullet$] We propose a new prompt learning paradigm \method~for CIL, where we adopt prompt addition to the CLS token instead of the conventional concatenation approach.
    \item[$\bullet$] Experiments show that \method~achieves state-of-the-art performance on popular continual learning benchmarks while offering: 1) lower inference overhead, 2) fewer trainable parameters, 3) simplified loss functions, and 4) reduced prompt-related hyperparameter optimization.
    \item[$\bullet$] Further experiments on various recognition tasks reveal that our method outperforms VPT in both performance and efficiency, positioning it as a promising candidate for a novel, general PEFT approach.
    
\end{itemize}

\section{Related Works}
\label{sec:related works}

\subsection{Prompt Tuning}
In recent years, parameter-efficient fine-tuning (PEFT) methods~\cite{he2021adapter, houlsby2019parameter, li2021prefix, hu2022lora, zaken2021bitfit, jia2022visual} have gained significant attention as alternatives to full model fine-tuning. These approaches aim to adapt large pre-trained models to downstream tasks by updating only a small subset of parameters, offering a more efficient and scalable solution.

Prompt tuning~\cite{lester2021power}, one of the most prominent PEFT approaches, was initially introduced in natural language processing. It began with discrete text templates designed to guide model behavior, which later evolved into continuous prompt tuning~\cite{li2021prefix}, where learnable vectors are optimized to better align with task-specific objectives. In the realm of computer vision, Visual Prompt Tuning (VPT)~\cite{jia2022visual} extended this concept to visual tasks by prepending learnable tokens to image embeddings in Vision Transformers. VPT demonstrated strong performance across a wide range of vision tasks, showcasing the versatility of prompt-based methods.

However, while effective, VPT-based approaches introduce certain limitations. The concatenation of prompts to the input sequence inevitably increases the computational overhead during inference, making them less efficient in practice. Furthermore, compared to their counterparts in NLP, concatenated prompt tokens remain largely opaque and difficult to interpret in the vision domain. As such, in this work, we propose a refined framework that rethinks how visual prompts should be designed and utilized.

\subsection{Class-Incremental Learning}

Class-incremental learning (CIL) aims to learn new classes sequentially while retaining knowledge of previously learned ones. A major challenge in CIL is catastrophic forgetting, where the model's performance on earlier classes deteriorates as it adapts to new ones. Traditional approaches to address this challenge fall into three main categories: rehearsal-based methods~\cite{rebuffi2017icarl, belouadah2019il2m}, architectural methods~\cite{rusu2016progressive,hung2019compacting,mallya2018piggyback}, and regularization-based methods~\cite{kirkpatrick2017overcoming, li2017learning}.

Recent advances in class-incremental learning (CIL) have shifted toward parameter-efficient solutions, particularly prompt-based approaches~\cite{wang2022l2p, wang2022dualprompt, chen2024promptfusion}. L2P~\cite{wang2022l2p} pioneered this direction by introducing a pool of learnable prompts dynamically selected for different tasks. Building on this foundation, DualPrompt~\cite{wang2022dualprompt} introduced the use of both general and task-specific prompts, while Coda-Prompt~\cite{smith2023coda} proposed leveraging a weighted average of all prompts in the pool to improve knowledge transfer. However, despite their effectiveness, these prompt-based methods often require an additional forward pass to query the prompt pool and rely on VPT-style prompt concatenation that extends the input sequence, which significantly increases computational costs during inference.

\section{Preliminaries}

\subsection{Class Incremental Learning} 
Our work focuses on class-incremental learning within the image classification domain. During training, images are presented sequentially, and at inference, the model is not provided with any information about the task identity. Formally, we consider training a model across $N$ tasks, $\mathcal{T} = (T_1, T_2, ..., T_N)$, where each task is associated with a dataset $\mathcal{D}(\mathcal{X}^t, \mathcal{Y}^t)$. Here $\mathcal{Y}^i \cap \mathcal{Y}^j = \emptyset$ for $i \neq j$.

\subsection{Visual Prompt Tuning} 
For a standard ViT with $L$ transformer layers, the input to each transformer layer can be represented as:
\begin{equation}
     X = [x_{cls}; x_1; ...; x_m],
\end{equation}
where $x_{cls} \in \mathbb{R}^d$ denotes the CLS token and the rest $x_i \in \mathbb{R}^d$ represents the embedded image tokens ($d$ is the embedding dimension). In VPT, learnable prompt tokens are inserted into the input sequence of the transformer layers. Depending on whether the framework is shallow or deep, these prompt tokens can either be concatenated only at the first transformer layer or at every transformer layer. Formally, let the learnable prompt tokens be denoted as $p_i \in \mathbb{R}^d$. The input is then reformulated as:
\begin{equation}
\label{eqn:vpt}
   X = [x_{cls}; p_1; ...; p_n; x_1; ...; x_m].
\end{equation}
During training, only the prompt tokens are updated, while all other model parameters remain frozen.

\subsection{Prompt-Based Continual Learning} 
In existing prompt-based continual learning frameworks, most methods maintain a prompt pool containing candidate prompts for each task. Each prompt $\bm{p}_i$ is associated with a learnable key $\bm{k}_i \in \mathbb{R}^d$. For a given input image $\bm{x}$, a query $q(\bm{x})$ is generated by passing the image through a raw ViT encoder: $q(\bm{x}) \in \mathbb{R}^d = \theta(\bm{x})$, where $\theta$ denotes the pretrained ViT encoder. Prompts are selected based on the similarity between their keys and the query, computed as $\gamma(q(\bm{x}), \bm{k}_i)$, where $\gamma$ represents cosine similarity. The selected prompts are then concatenated to the input sequence following \cref{eqn:vpt}. Clearly, this framework incurs additional computational costs due to the extra forward pass required to compute the query and a longer input sequence. Our proposed method addresses and improves upon these limitations, as will be introduced next.

\begin{figure*}[h]
  \centering
   \includegraphics[width=\linewidth]{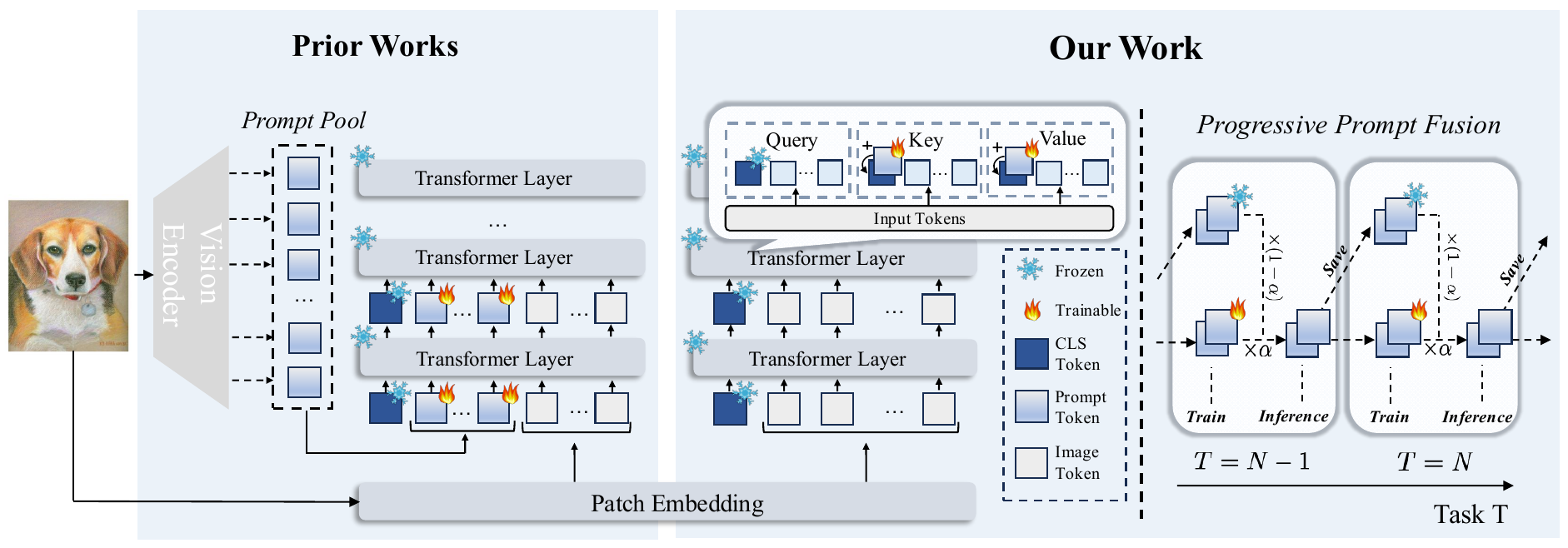}
   \caption{Overview of \method~compared to prior prompt-based methods. In our approach, each transformer layer is equipped with a pair of learnable key and value prompt tokens, which are directly added to the key and value vectors generated by the CLS token, respectively. This differs from previous prompt-based methods that typically maintain a pool of prompts and require an additional forward pass through the ViT encoder for prompt selection. Additionally, our method progressively fuses the learned prompts with previously stored ones during inference, enabling the model to retain performance on older tasks while learning new ones.}
   \label{fig:architecture}
\end{figure*}

\section{Method}
\label{sec:method}
The goal of this paper is to develop an efficient PEFT framework for continual learning that eliminates the computational overhead of traditional prompt-based methods while maintaining or improving performance. To this end, we propose a novel approach that combines prompt learning with additive attention modifications focused on the CLS token. In the following, we first introduce our core architectural design for prompt addition in \cref{sec:method_prompt_addition}, followed by the training objective of our approach in \cref{sec:method_train_inference}. Finally, we detail the proposed Progressive Prompt Fusion (PPF) inference strategy for mitigating catastrophic forgetting in \cref{sec:method_ema_prompt}

\subsection{Additive Prompt Design}
\label{sec:method_prompt_addition}

Different from previous approaches that concatenate visual prompt tokens with input embeddings, we propose directly modifying the attention computation through additive operations only on the CLS token. An overview of our architecture design and its comparison against previous prompt-based methods is illustrated in \cref{fig:architecture}. Our design is primarily motivated by two key insights: (1) the CLS token serves as a crucial aggregator of visual information in Vision Transformers~\cite{wang2024cls}, and (2) additive modifications can preserve the input sequence length and provide a more direct way to influence the model's feature extraction process compared to concatenation-based approaches.

In a standard Vision Transformer, given an input image $I$, it is first divided into non-overlapping patches and linearly projected to obtain a sequence of patch embeddings. A CLS token is prepended to this sequence that serves as a global representation for subsequent transformer layers. In our method, we enhance this process by introducing two learnable prompt vectors per layer, $p^l_k, p^l_v \in \mathbb{R}^d$, which are specifically designed to modify the key and value representations of the CLS token during self-attention. For each transformer layer $l$, we first extract the key vector $k^l_{cls}$ and value vector $v^l_{cls}$ of the CLS token, and then apply our additive modification:

\begin{equation}
    \hat{k}^l_{cls} = k^l_{cls} + p^l_k, \hat{v}^l_{cls} = v^l_{cls} + p^l_v
\end{equation}

\noindent with the key and value vectors of all patch tokens remaining unmodified. This is followed by the usual self-attention:

\begin{equation}
    \text{Attention}(Q^l, \hat{K}^l, \hat{V}^l) = \text{softmax}(\frac{Q^l (\hat{K}^l)^T}{\sqrt{d}})\hat{V}^l,
\end{equation}

\noindent where $\hat{K}^l$ and $\hat{V}^l$ are the modified keys and values matrices. 

\subsection{Training Objective}
\label{sec:method_train_inference}

Unlike previous prompt-based methods that often employ complex training objectives with multiple loss terms~\cite{smith2023coda, wang2023isolation, kurniawan2024evolving}, our approach maintains simplicity by utilizing \textbf{only the standard cross-entropy loss} for classification. This simplification is enabled by our efficient prompt design that eliminates the need for additional regularization or prompt selection mechanisms. Specifically, for a given task $t$, let $\mathcal{D}_t = \{(x_i, y_i)\}_{i=1}^{N_t}$ denote the training data, where $x_i$ represents input images and $y_i$ their corresponding labels. Our trainable parameters consist solely of the prompt vectors $P_t = \{p^l_{k,t}, p^l_{v,t}\}_{l=1}^L$ and a task-specific classification head $h_t$, while keeping the pre-trained transformer parameters frozen. The training objective is simply:

\begin{equation}
\label{eqn:loss}
    \mathcal{L} = -\frac{1}{N_t}\sum_{i=1}^{N_t} \log \frac{\exp(h_t(f(x_i))_{y_i})}{\sum_{j}\exp(h_t(f(x_i))_j)},
\end{equation}

\noindent where $f(x_i)$ represents the final CLS token embedding after applying our prompt-based modifications through all transformer layers.

\begin{table*}[!t]
\begin{center}
\resizebox{\linewidth}{!}{
\begin{tabular}{l|cc|cc|cc|cc}
\toprule 
\multirow{2}{*}{\textbf{Method}} & \multicolumn{2}{c|}{\textbf{Split CIFAR-100}}
 &  \multicolumn{2}{c|}{\textbf{Split ImageNet-R}}&  \multicolumn{2}{c|}{\textbf{Split CUB200}}&  \multicolumn{2}{c}{\textbf{Split Stanford Cars}}\\ 
& \multicolumn{1}{c}{Avg Acc($\uparrow$)} & \multicolumn{1}{c|}{Forgetting($\downarrow$)}   & \multicolumn{1}{c}{Avg Acc($\uparrow$)} & \multicolumn{1}{c|}{Forgetting($\downarrow$)}& \multicolumn{1}{c}{Avg Acc($\uparrow$)} & \multicolumn{1}{c|}{Forgetting($\downarrow$)} & \multicolumn{1}{c}{Avg Acc($\uparrow$)} & \multicolumn{1}{c}{Forgetting($\downarrow$)}\\
\midrule

 LAE~\cite{gao2023lae} & 86.09 \tsb{\( \pm \)0.55} & 6.22 \tsb{\( \pm \)0.17} & 73.88 \tsb{\( \pm \)0.43} & 5.08 \tsb{\( \pm \)0.33} & - & - & - & -  \\
 InfLoRA~\cite{liang2024inflora} & 86.91 \tsb{\( \pm \)0.34} & 4.22 \tsb{\( \pm \)0.37} & 76.13 \tsb{\( \pm \)0.44} & 5.22 \tsb{\( \pm \)0.48} & -& -& -& - \\
 EASE~\cite{zhou2024ease} & 87.28 \tsb{\( \pm \)0.71} & 4.81 \tsb{\( \pm \)0.42} & 75.82 \tsb{\( \pm \)0.54} & 5.83 \tsb{\( \pm \)0.37} & - & - & - & - \\
 
\midrule

L2P~\cite{wang2022l2p}  &83.21 \tsb{\( \pm \)0.06} & 8.81 \tsb{\( \pm \)0.32}& 72.64 \tsb{\( \pm \)0.70} & 4.22 \tsb{\( \pm \)0.49}  & 71.22 \tsb{\( \pm \)0.54}  & 10.68 \tsb{\( \pm \)0.61}  & 60.39 \tsb{\( \pm \)1.99}  & 13.00 \tsb{\( \pm \)0.12}  \\

DualPrompt~\cite{wang2022dualprompt} & 82.03 \tsb{\( \pm \)0.44} & 7.48 \tsb{\( \pm \)1.10} &  69.12 \tsb{\( \pm \)0.38}  & 4.69 \tsb{\( \pm \)0.44}  &71.55 \tsb{\( \pm \)0.73}  & 10.21 \tsb{\( \pm \)0.43} & 57.27 \tsb{\( \pm \)0.34}   & 16.31 \tsb{\( \pm \)0.97} \\

Coda-Prompt~\cite{smith2023coda} & 86.92 \tsb{\( \pm \)0.77}& 5.10 \tsb{\( \pm \)0.43}&   73.21 \tsb{\( \pm \)0.28}  & 5.69 \tsb{\( \pm \)0.28} & 73.25 \tsb{\( \pm \)0.71} & 11.02 \tsb{\( \pm \)0.88}  &62.24 \tsb{\( \pm \)0.14}  & 15.08 \tsb{\( \pm \)0.89} \\

ESN~\cite{wang2023isolation} & 86.42 \tsb{\( \pm \)0.80} & 6.08 \tsb{\( \pm \)0.48} & 75.11 \tsb{\( \pm \)0.36}  & 5.68 \tsb{\( \pm \)0.77}  & 71.20 \tsb{\( \pm \)0.54} & 9.82 \tsb{\( \pm \)0.43} &  56.91 \tsb{\( \pm \)0.56}  & 13.50 \tsb{\( \pm \)1.64} \\

EvoPrompt~\cite{kurniawan2024evolving} & 87.97 \tsb{\( \pm \)0.30} & 3.12 \tsb{\( \pm \)0.50} &  76.83 \tsb{\( \pm \)0.08}    & 3.34 \tsb{\( \pm \)0.07}  & - & - & - &  -\\

OVOR-Deep~\cite{huang2024ovor} & 85.99 \tsb{\( \pm \)0.89} & 6.42 \tsb{\( \pm \)2.03}&  76.11 \tsb{\( \pm \)0.21}    & 7.16 \tsb{\( \pm \)0.34} & 78.11 \tsb{\( \pm \)0.47}   & 7.95 \tsb{\( \pm \)0.77}  & 47.59 \tsb{\( \pm \)3.12}   & 15.66 \tsb{\( \pm \)1.46} \\

CPrompt~\cite{gao2024consistent} & 87.82 \tsb{\( \pm \)0.21} & 5.06 \tsb{\( \pm \)0.50}& 77.14 \tsb{\( \pm \)0.11}   & 5.97 \tsb{\( \pm \)0.68}  & 77.09  \tsb{\( \pm \)0.64}& 10.27 \tsb{\( \pm \)0.54} & 66.77 \tsb{\( \pm \)0.37}  & 13.95 \tsb{\( \pm \)0.46} \\

\midrule

\method~(Ours) & \textbf{88.88 \tsb{\( \pm \)0.65}} & 3.47 \tsb{\( \pm \)0.55} & \textbf{79.40 \tsb{\( \pm \)0.47} } & 4.38 \tsb{\( \pm \)0.46} & \textbf{78.50 \tsb{\( \pm \)0.94} } & 7.76 \tsb{\( \pm \)0.88} & \textbf{71.04 \tsb{\( \pm \)0.38}}  & 7.95 \tsb{\( \pm \)0.69} \\

\bottomrule
\end{tabular}}
\caption{Performance comparisons with both non-prompt and prompt-based methods on Split CIFAR-100, Split ImageNet-R, Split CUB200, and Split Stanford Cars. The experiments are conducted under a 10-task continual learning setting. Best results are marked in \textbf{bold}.} 
\label{table:main_results}
\end{center}
\end{table*}
\subsection{Progressive Prompt Fusion Inference Strategy}
\label{sec:method_ema_prompt}

To effectively mitigate catastrophic forgetting while maintaining performance on new tasks, we adopt a simple Progressive Prompt Fusion (PPF) strategy for prompt adaptation during inference. This approach ensures a smooth transition between tasks while preserving knowledge from previously learned tasks.

Formally, let $P_t = \{p^l_{k,t}, p^l_{v,t}\}_{l=1}^L$ denote the set of all prompt vectors after training on task $t$, where $L$ is the number of transformer layers. Before training on a new task $t+1$, we store the current prompts $P_t$. After completing the training on task $t+1$, instead of directly using the newly trained prompts $P_{t+1}$, we compute a weighted average between the old and new prompts:

\begin{equation}
\label{eqn:ema}
    P^{\text{PPF}}_{t+1} = \alpha P_t + (1-\alpha) P_{t+1},
\end{equation}

\noindent where $\alpha \in [0,1]$ is a hyperparameter controlling the balance between retaining old knowledge and adapting to new tasks. Note that this PPF strategy is applied exclusively during inference. During training, the model only uses the current prompt vectors to ensure effective learning of task-specific features. Once adjusted, the PPF prompts are used for all subsequent inferences until the next task is encountered. 
\section{Experiments}
\subsection{Experimental Setup}
\noindent\textbf{Datasets.} We evaluate our method on four widely used benchmark datasets for continual learning: Split CIFAR-100~\cite{krizhevsky2009cifar}, Split ImageNet-R~\cite{hendrycks2021imgnetr}, Split CUB200~\cite{wah2011cub}, and Split Stanford Cars~\cite{gebru2017cars}. CIFAR-100 is a standard image classification dataset comprising 60,000 images of size 32 $\times$ 32 pixels. It includes 100 classes, each with 600 images (500 for training and 100 for testing). ImageNet-R is a challenging benchmark designed to evaluate model robustness to natural distribution shifts. It contains 30,000 images across 200 ImageNet classes, featuring diverse artistic renditions such as cartoons, graffiti, origami, and more. The combination of both semantic and covariate shifts makes ImageNet-R one of the most difficult datasets for continual learning. CUB200 and Stanford Cars, on the other hand, are fine-grained visual classification datasets. CUB200 consists of 11,788 images spanning 200 bird species, while Stanford Cars includes 16,185 images across 196 car classes categorized by make, model, and year.

\noindent\textbf{Evaluation metrics.} In this work, we mainly use the conventional metrics Average Accuracy and Forgetting. Here, Average Accuracy reflects the overall performance of the model on all tasks after finished training on all tasks, and Forgetting measures the degree of performance decline on previously learned tasks. Formally, let $R_{t, i}$ be the classification accuracy of task $T_i$ after training on task $T_t$, then Average Accuracy $\mathcal{A}_B$ for task $T_T$ is defined as $\mathcal{A}_B = \frac{1}{T}\sum_{i=1}^{T} R_{T, i}$, and Forgetting is defined as $\mathcal{F} = \frac{1}{T-1} \sum_{i=1}^{T-1}(R_{i,i}-R_{T,i})$.

\noindent\textbf{Implementation details.}  Following previous works, our method employs a ViT-B/16 architecture pretrained on ImageNet-21k. For all experiments, we apply the Adam optimizer with a batch size of 64. The learning rate is set to 0.004, 0.003, 0.02 and 0.005 for Split CIFAR-100, Split ImageNet-R, Split CUB200 and Split Stanford Cars, respectively. We train 18 epochs for Split CUB200, 30 epochs for both Split CIFAR-100 and Split ImageNet-R, and 50 epochs for Split Stanford Cars. As illustrated in \cref{sec:method}, the only additional hyperparameter introduced in \method~is the PPF coefficient $\alpha$ in \cref{eqn:ema}. It is set to 0.8 for Split ImageNet-R and 0.7 otherwise. We apply \method~to every transformer layer due to its lightweight design. 

For the main experiments, all tested datasets are split into 10 tasks. No memory is used throughout the training process. During inference, no task identity is provided to the model. We follow the convention of other prompt-based approaches and primarily compare \method~against them, including L2P~\cite{wang2022l2p}, DualPrompt~\cite{wang2022dualprompt}, Coda-Prompt~\cite{smith2023coda}, ESN~\cite{wang2023isolation}, EvoPrompt~\cite{kurniawan2024evolving}, OVOR~\cite{huang2024ovor}, and CPrompt~\cite{gao2024consistent}. Additionally, we compare \method~with a few of the latest non-prompt-based methods, including LAE~\cite{gao2023lae}, InfLoRA~\cite{liang2024inflora}, and Ease~\cite{zhou2024ease}, though this is not our primary focus.

\begin{table*}[h]
\begin{minipage}{0.48\textwidth}
\begin{center}
\setlength{\tabcolsep}{4mm}
\resizebox{0.9\linewidth}{!}{
\begin{tabular}{l|cc}
\toprule 
\multirow{2}{*}{\textbf{Method}}  & \multirow{2}{*}{GFLOPs}  & Trainable Prompt \\
 & & Parameters (M)\\

\midrule

L2P & 35.73 (2.13$\times$) &  0.04 (2.0$\times$)\\
DualPrompt & 33.69 (2.01$\times$) & 0.32 (16.0$\times$)\\
Coda-Prompt & 33.67 (2.00$\times$) & 3.07 (153.5$\times$) \\
ESN & 25.36 (1.51$\times$)& 0.08 (4.0$\times$)\\
EvoPrompt & 36.37 (2.16$\times$)& 0.04 (2.0$\times$)\\
OVOR-Deep & 16.81 (1.00$\times$)& 0.11 (5.5$\times$)\\
CPrompt & 37.87 (2.25$\times$)& 0.77 (38.5$\times$)\\

\midrule

\method~(Ours) & \textbf{16.80 (1.00$\times$)} & \textbf{0.02 (1.0$\times$)}\\

\bottomrule
\end{tabular}}
\caption{GFLOPs and number of trainable prompt parameters on Split ImageNet-R. For GFLOPs, we mainly focus on the self-attention blocks. We also report the respective ratio to \method.} 
\label{table:efficiency}
\end{center}
\end{minipage}
\hfill
\begin{minipage}{0.48\textwidth}
\centering
\resizebox{\linewidth}{!}{
\begin{tabular}{l|cc|cc}
\toprule 
\multirow{2}{*}{\textbf{Method}} & \multicolumn{2}{c|}{\textbf{20-Task Split CIFAR-100}}
 &  \multicolumn{2}{c}{\textbf{20-Task Split ImageNet-R}}\\ 
& \multicolumn{1}{c}{Avg Acc($\uparrow$)} & \multicolumn{1}{c|}{Forgetting($\downarrow$)}   & \multicolumn{1}{c}{Avg Acc($\uparrow$)} & \multicolumn{1}{c}{Forgetting($\downarrow$)}\\

\midrule
Upper Bound & 92.83 & -& 84.07 & -  \\
\midrule

L2P & 78.10 \tsb{\( \pm \)0.72} & 10.04 \tsb{\( \pm \)0.52}& 69.94 \tsb{\( \pm \)1.03} & 5.46 \tsb{\( \pm \)0.83}    \\

DualPrompt & 77.12 \tsb{\( \pm \)0.72} & 8.13 \tsb{\( \pm \)1.03} &  66.03 \tsb{\( \pm \)1.06}  & 5.92 \tsb{\( \pm \)0.35}   \\

Coda-Prompt & 79.58 \tsb{\( \pm \)0.79}& 5.69 \tsb{\( \pm \)0.62}&   70.52 \tsb{\( \pm \)0.68}  & 5.63 \tsb{\( \pm \)0.61} \\

ESN & 80.56 \tsb{\( \pm \)0.94} & 6.08 \tsb{\( \pm \)0.48} & 70.57 \tsb{\( \pm \)0.62}  & 6.84 \tsb{\( \pm \)0.36}   \\

EvoPrompt & 84.64 \tsb{\( \pm \)0.64} & 4.78 \tsb{\( \pm \)0.29} &  74.68 \tsb{\( \pm \)0.51}    & 3.24 \tsb{\( \pm \)0.23} \\

OVOR-Deep & 84.10 \tsb{\( \pm \)0.21} & 5.97 \tsb{\( \pm \)0.86}&  72.36 \tsb{\( \pm \)0.45}    & 5.65 \tsb{\( \pm \)0.49}  \\

CPrompt & 84.93 \tsb{\( \pm \)0.36}& 5.80 \tsb{\( \pm \)0.65}   & 74.46 \tsb{\( \pm \)0.33} & 6.41 \tsb{\( \pm \)0.13} \\

\midrule

\method~(Ours) & \textbf{86.37 \tsb{\( \pm \)0.46}} &  4.57 \tsb{\( \pm \)0.88} & \textbf{76.22 \tsb{\( \pm \)0.04} } & 6.91\tsb{\( \pm \)0.33}\\

\bottomrule
\end{tabular}}
\captionof{table}{Average accuracy and Forgetting results of \method~on Split CIFAR-100 and Split ImageNet-R under a longer 20-task continual learning setting.} 
\label{table:long_task}
\end{minipage}
\end{table*}

\subsection{Performance Comparison against Other Approaches}

The results of \method~on the four evaluated datasets are presented in \cref{table:main_results}. For Split CIFAR-100, \method~achieves the highest average accuracy. Compared to the second-best performing method, EvoPrompt, our approach surpasses it by 0.91\%. Since CIFAR-100 is a relatively small and less complex dataset, we believe this improvement is still significant, highlighting the effectiveness of our method. For Split ImageNet-R, one of the most challenging datasets for continual learning, \method~outperforms all other methods by over 2\%, demonstrating its robustness in tackling complex and diverse classification tasks. On the two fine-grained datasets, Split CUB200 and Split Stanford Cars, \method~still achieves the best performance compared to other approaches. On average, our method outperforms them by 4.76\% and 12.51\%, respectively. Notably, for Split Stanford Cars, \method~exceeds the second-best method, CPrompt, by a substantial margin of 4.27\%. This significant improvement underscores the strength of our approach in capturing subtle inter-class variations, which is crucial for fine-grained classification tasks. All these results provide compelling evidence of the effectiveness of \method~across a diverse range of datasets, from general to fine-grained, and from relatively simple to highly challenging tasks.

\begin{figure*}[t]
  \centering
   \includegraphics[width=0.9\linewidth]{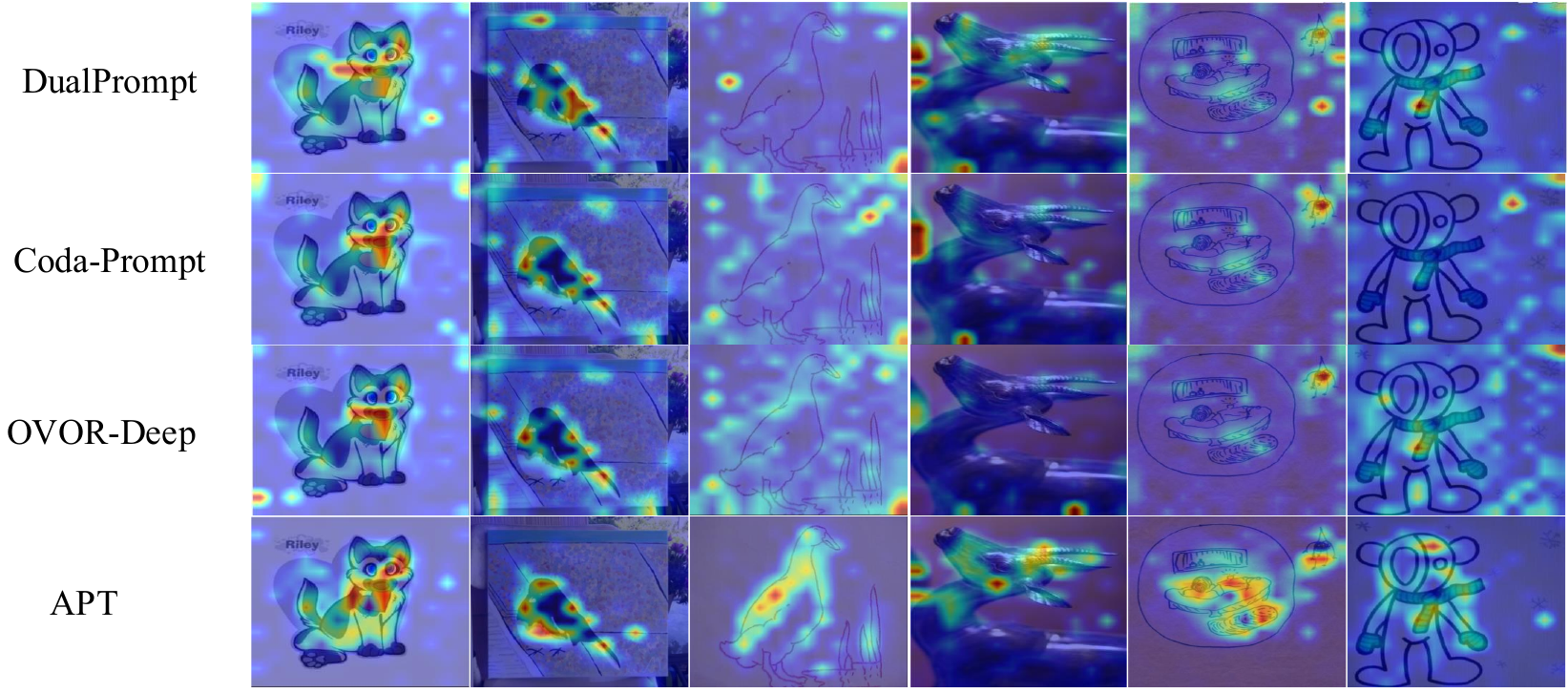}
   \caption{Attention heatmap visualization comparing prior methods and ours, highlighting the improved attention achieved by \method.}
   \vspace{1ex}
   \label{fig:attention}
\end{figure*}

\subsection{Efficiency Comparison with Prompt-Based Methods}
In addition to the classification accuracy, we compare the computational efficiency and parameter overhead of our method with other prompt-based continual learning approaches on the Split ImageNet-R dataset. The results are shown in \cref{table:efficiency}. The comparison is based on two key metrics: the total GFLOPs mainly associated with the self-attention blocks, and the number of trainable prompt parameters. 

Clearly, our method achieves the lowest inference cost among all compared methods, requiring only 16.80 GFLOPs, which is almost the same as a plain ViT. While this matches the GFLOPs of OVOR-Deep, results from \cref{table:main_results} show that we consistently outperform it in terms of accuracy across all four datasets. In contrast, other methods such as L2P, DualPrompt, EvoPrompt, and CPrompt require up to 2.25 $\times$ more GFLOPs than our approach, primarily due to the query mechanism that requires an additional forward pass through the ViT encoder. This highlights the computational efficiency of our design.
 
When it comes to trainable prompt parameters, our method significantly reduces parameter overhead, requiring only 0.02 million trainable parameters, the smallest among all methods. This represents at least a 50\% reduction in trainable parameters compared to other baselines. For Coda-Prompt and CPrompt, the reduction is even more dramatic, reaching 99.3\% and 97.4\%, respectively. Compared to OVOR-Deep, which has a similar inference cost to \method, our approach reduces the number of trainable parameters by a factor of 5.5, demonstrating a clear advantage in parameter efficiency . It is worth noting that this comparison accounts only for parameters exclusive to prompt tokens. Most of the methods also require a learnable key paired with each prompt token for training. Consequently, when considering the total number of trainable parameters, our method achieves even greater reductions in parameter overhead.

Overall, our approach not only achieves the best performance in terms of accuracy but also sets a new standard for efficiency, requiring the least amount of computational resources among all prompt-based continual learning methods. This makes our method a strong candidate for setting up a new paradigm for prompt-based CIL approaches.

\subsection{Attention Visualization}
To further evaluate \method~against other prompt-based methods, we visualize their attention heatmaps in \cref{fig:attention}. Notably, while prior approaches often exhibit scattered or diffuse attention, APT exhibits highly focused attention that consistently emphasizes the most critical regions of the input images. This demonstrates that our proposed prompting paradigm more effectively captures task-relevant features, enabling improved feature extraction and decision-making.

\subsection{Long Task Sequence}
To further assess the effectiveness and robustness of \method, we evaluate its performance in a longer task sequence setting. Specifically, we test it on Split CIFAR-100 and Split ImageNet-R with 20 tasks. The detailed results are presented in \cref{table:long_task}. As shown, \method~consistently achieves the best performance among all compared methods, with 86.37\% accuracy on Split CIFAR-100 and 76.22\% on Split ImageNet-R. On average, this represents a performance boost of 5.08\% and 5.00\%, respectively. These results demonstrate that \method~maintains high accuracy and stability throughout extended task sequences, showcasing its scalability and effectiveness in more challenging, longer task settings.

\begin{table}[h]
\setlength{\tabcolsep}{1mm}
\resizebox{\linewidth}{!}{
\begin{tabular}{l|cc|cc}
\toprule 
\multirow{2}{*}{\textbf{Method}} & \multicolumn{2}{c|}{\textbf{Split CIFAR-100}}
 &  \multicolumn{2}{c}{\textbf{Split ImageNet-R}}\\ 
& \multicolumn{1}{c}{Avg Acc($\uparrow$)} & \multicolumn{1}{c|}{Forgetting($\downarrow$)}   & \multicolumn{1}{c}{Avg Acc($\uparrow$)} & \multicolumn{1}{c}{Forgetting($\downarrow$)}\\

\midrule

w/o PPF & 88.17 \tsb{\( \pm \)0.51} & 7.05 \tsb{\( \pm \)0.22}& 78.56 \tsb{\( \pm \)0.45} & 8.83 \tsb{\( \pm \)0.29}    \\
w/o add KV & 87.93 \tsb{\( \pm \)0.45} & 5.24 \tsb{\( \pm \)0.43} &  78.31 \tsb{\( \pm \)0.55}  & 5.18 \tsb{\( \pm \)0.36}\\
\method~(Ours) & 88.88 \tsb{\( \pm \)0.65} & 3.47 \tsb{\( \pm \)0.55} &  79.40 \tsb{\( \pm \)0.47}  & 4.38 \tsb{\( \pm \)0.46}   \\

\bottomrule
\end{tabular}}
\captionof{table}{Ablation study on the PPF inference strategy and prompt additive design choice on Split CIFAR-100 and Split ImageNet-R.} 
\label{tab:ablation}
\end{table}

\subsection{Ablation and Hyperparameter Analysis}
We present the results of an ablation study on our PPF strategy in \cref{tab:ablation}. As demonstrated, incorporating the PPF strategy consistently improves the Average Accuracy. More importantly, it significantly mitigates Forgetting, with substantial reductions observed on both Split CIFAR-100 (3.58\%) and Split ImageNet-R (4.45\%). Additionally, we perform an ablation study on the design choice of adding prompts to the KV vectors of the CLS token. Specifically, we evaluate the alternative of directly adding prompts to the CLS token at the input level. Once again, the results in \cref{tab:ablation} confirm consistent improvements in Average Accuracy.

\begin{table*}[!t]
  \centering
  \setlength{\tabcolsep}{2mm}
  \resizebox{0.9\linewidth}{!}{
  \begin{tabular}{lccccccccccccccc} \toprule
\multirow{2}{*}{\textbf{Method}}     & \multirow{2}{*}{CUB200}  & \multirow{2}{*}{Flowers102} & Stanford  & Stanford  & \multirow{2}{*}{SVHN} & \multirow{2}{*}{DTD} & Patch &\multirow{2}{*}{EuroSat}& \multirow{2}{*}{Avg} \\
   &&&Cars& Dogs & & & Camelyon \\ 
    \midrule
    Full Fine-tuning &87.3& 98.8 &84.5& 89.4& 87.4 &64.3 &79.7 & 95.7 & 85.9 \\
    
    \midrule
    Linear Probing&85.3& 97.9& 51.3& 86.2 & 36.6 &63.2 &78.5 &87.5 & 73.3\\
    BitFit & 88.4& 98.8& 79.4 &91.2& 59.9 &59.2&78.7&91.6& 80.9\\
    Adapter-64 &87.1 &98.5 &68.6 &89.8 & 36.3 & 62.7& 76.3& 87.5 & 75.9\\
    Adapter-256 &87.2& 98.5 &68.6& 89.9 &34.6 &63.2 & 76.3 &88.0 &75.8\\
    VPT-shallow  & 86.7 & 98.4 & 68.7 & 90.7 & 74.5 & 62.6 & 78.2 & 92.0 & 81.5 \\

    VPT-deep  &
    88.5 & 98.9& 75.2 & \textbf{91.1} & 78.1 &65.8 & \textbf{81.8} & \textbf{96.1} & 84.4\\ 

    \midrule

     \method~(Ours) & \textbf{89.1}& \textbf{99.1} & \textbf{84.2} & 90.9 &\textbf{82.9} & \textbf{68.1} & 79.9 & \textbf{96.1} & \textbf{86.3}\\
    
    \bottomrule
  \end{tabular}}
    \caption{Performance comparison on general recognition datasets outside the continual learning scenario.}
    \vspace{2ex}
  \label{tab:general_recognition}
\end{table*}

\begin{table*}[!t]
  \centering
  \setlength{\tabcolsep}{1mm}
  \resizebox{\linewidth}{!}{
  \begin{tabular}{lcc|cc|cc|cc|cc|cc} \toprule
\multirow{2}{*}{\textbf{Method}}     & \multicolumn{2}{c|}{CUB200}  & \multicolumn{2}{c|}{Stanford Cars}& \multicolumn{2}{c|}{Stanford Dogs}& \multicolumn{2}{c|}{SVHN} & \multicolumn{2}{c|}{Patch Camelyon} & \multicolumn{2}{c}{EuroSat} \\
 &GFLOPs & \# Param. &GFLOPs & \# Param. &GFLOPs & \# Param. &GFLOPs & \# Param. &GFLOPs & \# Param. &GFLOPs & \# Param. \\ 
    \midrule
    Full Fine-tuning & 16.80 & -  & 16.80 & - & 16.80 & - & 16.80 & - & 16.80 & - & 16.80 & - \\
    \midrule
    VPT-shallow  & 25.36 & 0.077 & 25.36 & 0.077 & 25.36 & 0.077 & 33.96 & 0.077 & 17.22 & 0.004 & 21.07 & 0.038\\

    VPT-deep  & 17.65 & 0.008 & 33.96 & 0.077 & 25.36 & 0.077 & 21.07 & 0.038 & 25.36 & 0.077 & 25.36 & 0.077 \\ 

    \midrule

     \method~(Ours) & 16.80 & 0.018  & 16.80 & 0.018 & 16.80 & 0.018 & 16.80 & 0.018 & 16.80 & 0.018 & 16.80 & 0.018\\
    
    \bottomrule
  \end{tabular}}
    \caption{GFLOPs and prompt parameter comparison between \method~and VPT.}
  \label{tab:general_efficiency}
  \vspace{2ex}
\end{table*}

\begin{figure}
    \centering
    \includegraphics[width=0.8\linewidth]{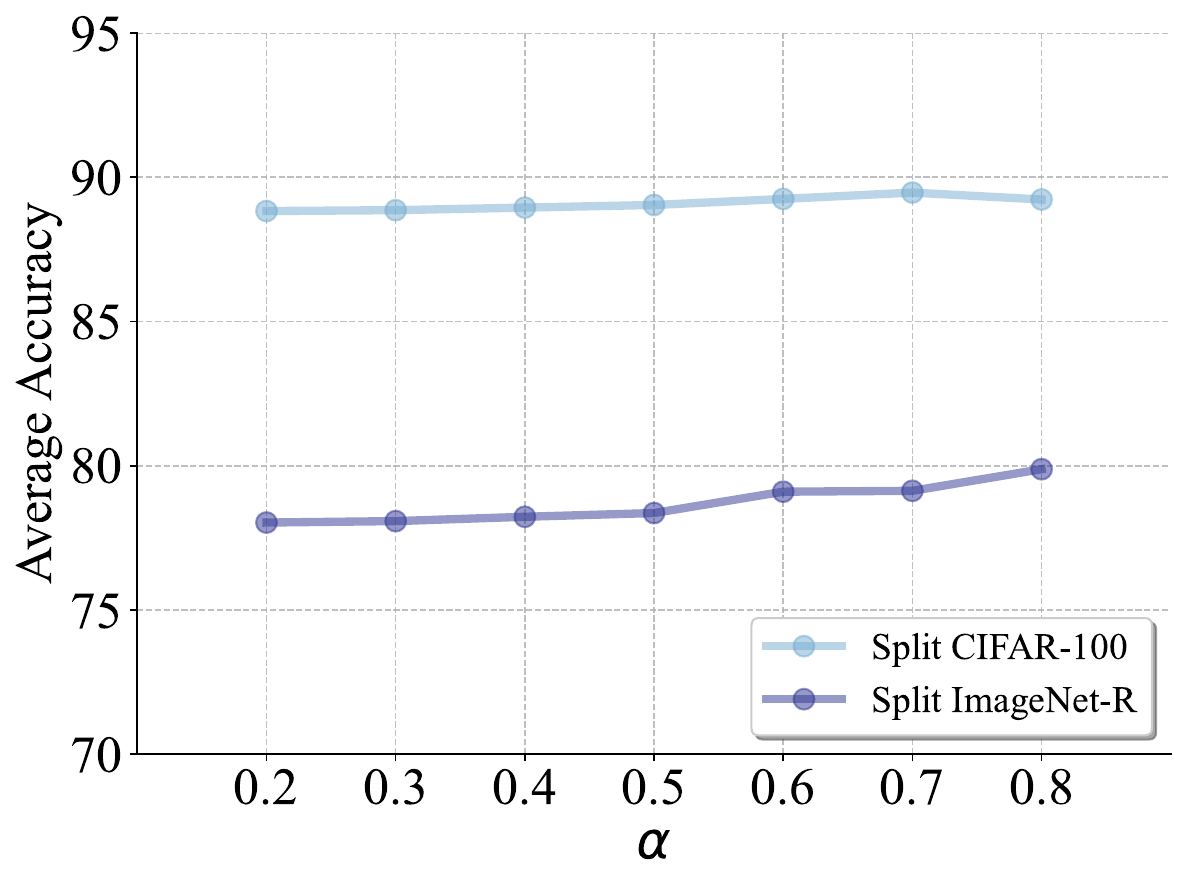}
    \caption{Hyperparameter analysis results on $\alpha$ from \cref{eqn:ema}}.
    \vspace{-2ex}
    \label{fig:hyperparameter_analysis}
\end{figure}

We further conduct a sensitivity analysis on the hyperparameter $\alpha$ in \cref{eqn:ema} using the Split CIFAR-100 and Split ImageNet-R datasets still under the 10-task setting. Specifically, we test $\alpha$ values ranging from 0.2 to 0.8. As clearly shown in \cref{fig:hyperparameter_analysis}, the performance of \method~exhibits relatively small fluctuations across different $\alpha$ values, highlighting the robustness of the PPF design.

Here we emphasize again that this is the only additional hyperparameter for \method, making it highly efficient in terms of hyperparameter optimization. Unlike other approaches, \method~eliminates the need to determine the optimal prompt length, as it is fixed at 2 per layer. For the selection of layers where \method~is applied, we empirically choose to apply it to all layers due to its extremely lightweight design, both in terms of trainable parameters and inference computation. Consequently, our approach requires \textbf{no tedious prompt-specific optimization}, significantly simplifying its practical use. This represents a major advancement over existing prompt-based methods. 

\subsection{General Recognition Tasks} 
Although \method~is mainly designed for CIL tasks, its architecture naturally positions it as a promising general-purpose PEFT approach. To validate this, we apply \method~to a broad range of downstream recognition tasks beyond the continual learning scenario. Following VPT, we evaluate \method~on 4 FGVC (Fine-Grained Visual Classification) datasets and 4 VTAB (Visual Task Adaptation Benchmark) datasets. The FGVC datasets focus on fine-grained categorization of visually similar objects, providing a benchmark to assess models’ ability to distinguish subtle inter-class differences. In contrast, the VTAB datasets encompass diverse tasks, spanning natural images, specialized domains, and structured scenes. The FGVC datasets we use in this section include CUB200~\cite{wah2011cub}, Oxford Flowers102~\cite{nilsback2008flowers}, Stanford Cars~\cite{gebru2017cars}, and Stanford Dogs~\cite{khosla2011dogs}, while the VTAB datasets~\cite{zhai2019vtab} we use are SVHN, DTD, Patch Camelyon, and EuroSat.

We compare \method~with other PEFT methods, with a particular focus on VPT as our direct competitor, and the results are shown in \cref{tab:general_recognition}. Our method clearly outperforms other approaches on most of the tested datasets, achieving an average performance improvement of 1.9\% over VPT-deep. Notably, \method~even surpasses full fine-tuning by 0.4\%, demonstrating its effectiveness. Additionally, we present the computational efficiency comparison in \cref{tab:general_efficiency}, reporting GFLOPs (focused on self-attention blocks) and the total number of prompt parameters. As shown, \method~consistently achieves the lowest GFLOPs at 16.8 across all datasets, with the prompt parameters fixed at 0.018M. In contrast, VPT-shallow and VPT-deep incur significantly higher computational costs and demonstrate strong dependence on the specific dataset being tested. Notably, for certain datasets (e.g., Stanford Cars and SVHN), the number of required prompt tokens for VPT reaches as high as 200, even exceeding the number of image tokens in the experimented ViT architecture (which is 196). Although this direction is not the primary focus of the paper, these results highlight \method's ability to deliver superior performance while maintaining significantly higher computational efficiency and scalability across diverse general recognition tasks. This underscores its strong potential as an effective and versatile PEFT method for future research.
\section{Conclusion}
\label{sec:conclusion}
In this study, we addressed the critical challenge of high computational costs in existing prompt-based class-incremental learning approaches. To tackle this, we proposed a novel prompt-tuning paradigm that replaced the conventional concatenation operation with an additive approach. Specifically, this addition is applied only to the key and value vectors generated by the CLS token during self-attention. Experiments on widely-used CIL benchmarks demonstrated that our method not only achieved state-of-the-art performance but also significantly reduced both inference costs and the number of trainable parameters. Furthermore, by evaluating our approach across a diverse range of recognition datasets, we showcased its potential as an effective and versatile general-purpose PEFT method.

\paragraph{\noindent Acknowledgement} This work was supported in part by the National Natural Science Foundation of China (Grant 62472098).  This work was supported by the Science and Technology Commission of Shanghai Municipality (No. 24511103100).

{
    \small
    \bibliographystyle{ieeenat_fullname}
    \bibliography{main}
}

\end{document}